\documentclass[conference]{IEEEtran}
\IEEEoverridecommandlockouts

\usepackage{cite}
\usepackage{amsmath,amssymb,amsfonts}
\usepackage{algorithmic}
\usepackage{graphicx}
\usepackage{textcomp}
\usepackage{xcolor}
\usepackage{hyperref}
\usepackage[caption=false]{subfig}
\usepackage{caption}
\usepackage{cuted}
\usepackage{booktabs}

\def\BibTeX{{\rm B\kern-.05em{\sc i\kern-.025em b}\kern-.08em
    T\kern-.1667em\lower.7ex\hbox{E}\kern-.125emX}}

\begin{document}

\title{TimeLineXR: An Interactive Exploration of Time in Vision-Language Models}

\author{%
\IEEEauthorblockN{Nidham Tekaya\IEEEauthorrefmark{1}\thanks{\IEEEauthorrefmark{1}Corresponding author.}}
\IEEEauthorblockA{\textit{Univ. of Applied Sciences St. Pölten}\\
\textit{TU Wien}\\
St. Pölten / Vienna, Austria\\
nidham.tekaya@ustp.at}
\and
\IEEEauthorblockN{Markus Passecker}
\IEEEauthorblockA{\textit{Univ. of Applied Sciences St. Pölten}\\
\textit{TU Wien}\\
St. Pölten / Vienna, Austria\\
markus.passecker@ustp.at}
\and
\IEEEauthorblockN{Matthias Zeppelzauer}
\IEEEauthorblockA{\textit{Univ. of Applied Sciences St. Pölten}\\
St. Pölten, Austria\\
matthias.zeppelzauer@ustp.at}
}

\maketitle

\begin{strip}
\centering
\includegraphics[width=0.95\textwidth]{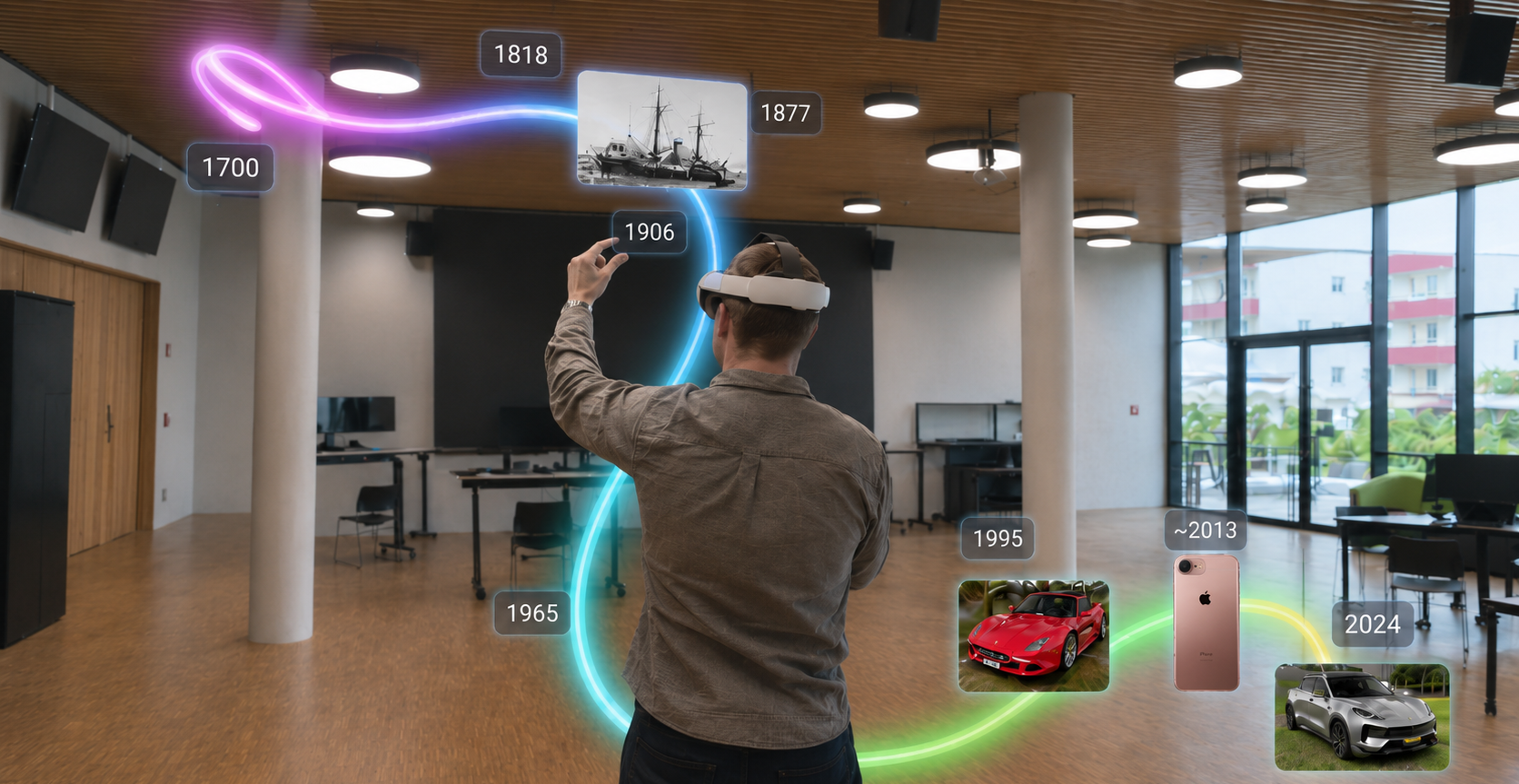}
\captionof{figure}{TimeLineXR in mixed reality. A user explores a vision-language
model's time axis at room scale; when an image card is thrown into the scene, the
system places it at its estimated year of first appearance on the time axis. The
supplementary video shows this XR workflow and the corresponding browser workflow.}
\label{fig:teaser}
\end{strip}

\begin{abstract}
Recently, it has been shown that vision-language models (VLMs) have a certain intrinsic understanding of time. They can estimate the temporal origin of objects and encode time along a low-dimensional, non-linear manifold inside their embedding space~\cite{tekaya2025matter}. That result is quantitative and reproducible, but the structure itself stays out of reach for end users: they cannot see it, query it with their own images, or inspect where a query lands relative to other years on the time axis. We present \textit{TimeLineXR}, an interactive demonstration that turns this time axis into an explorable 3D scene. A web demo lets users upload or select images and see each placed on a color-coded 3D time axis; a Three.js WebXR scene renders the same time axis as a room-scale representation that can be grabbed, scaled, inspected, and queried by throwing image cards into it. Both clients run the same prediction pipeline against the same cached temporal index: encode the query with CLIP, project it through cosine-kernel KPCA, and snap it to the B\'{e}zier time axis. The system turns each image into a point on the time axis, so users can compare where different objects land and read off their estimated first year of appearance.Code source and reproduction steps can be found in: \url{https://github.com/sterngefeuert/timeline-xr}.
\end{abstract}


\section{Introduction}

Temporal reasoning in images asks when an object or visual artifact first appeared
in time. In information retrieval terms, this is a query-by-example task over a
temporal index: an image query should be mapped to a year coordinate that can support
search, filtering, browsing, and comparison in multimedia collections.
 
Recent work has shown that VLMs implicitly encode time along a structured,
low-dimensional axis inside their joint image-text embedding
space~\cite{tekaya2025matter}. This \emph{time axis} acts as an internal timeline:
images of objects from similar periods cluster together, and the axis progresses
chronologically from older to more recent artifacts. This is an intriguing property
that may reveal something fundamental about the expressiveness and temporal awareness
of VLMs. Yet the time axis itself has not been explored interactively: users cannot
see it, walk along it, or query it with their own images to understand where a given
artifact lands and why. Enabling this exploration is a step toward interpretability
of VLM representations, which is a topic of growing interest in multimedia
analysis~\cite{smilkov2016embedding, marriott2018immersive}.
 
To enable this exploration, we introduce TimeLineXR. The system estimates each query
image's year of first appearance and places it on a browsable 3D representation of
the VLM's time axis. The web frontend opens this view in any browser; the WebXR
frontend adds depth, real-world scale, and hands-on manipulation. Seeing where images
land on the time axis, and being able to move around it, compare placements, and
inspect the structure, gives users direct intuition about how the model organises time
and where its estimates agree or diverge with expectations.

As supplementary material we include a demo video that walks through the headset
exploration and estimation workflow; upon acceptance, the WebXR application will be
hosted at a public HTTPS endpoint and the Gradio web demo will be released as a
Hugging Face Space.

This demonstration makes three contributions:
\begin{itemize}
    \item \textbf{A browser-accessible web demo} that lets users upload or select
          images and see their estimated first-appearance years rendered as placements
          on an interactive 3D time axis.
    \item \textbf{A zero-install WebXR interface} that renders the same time axis
          immersively at room scale, with grab, rotate, and scale manipulation and
          throw-to-time queries.
    \item \textbf{A single shared backend} serving both clients, which packages the
          time estimation method of~\cite{tekaya2025matter} (CLIP encoder,
          cosine-kernel KPCA projection, B\'{e}zier time axis, nearest-point
          assignment) for two deployment modes: embedded in a server-side Gradio Space
          for the web demo and exposed through a FastAPI HTTP service for the WebXR
          client.
\end{itemize}

\section{Related Work}

\textbf{Time-aware vision-language models.}
The reference method~\cite{tekaya2025matter} uses
CLIP\,ViT-B/32~\cite{radford2021clip} and EVA-CLIP~\cite{sun2023evaclip} as
representative backbones and shows that their time axis can be approximated
by a 1D B\'{e}zier curve in a kernel PCA subspace. Earlier visual time-estimation work
focused on dating historical photographs in a single visual
domain~\cite{barancova2023blind} and did not expose the underlying temporal
structure of the model.

\textbf{Embedding-space and retrieval interfaces.}
Embedding Projector~\cite{smilkov2016embedding} popularised interactive visualization of high-dimensional embeddings, but it is a generic projection viewer: it does not provide a temporal index or time estimation capabilities. \textit{TimeLineXR} instead treats the time axis itself as an explorable structure that users can query with their own images, see their placement on, and browse along.

\textbf{Immersive analytics and cultural heritage.}
Immersive analytics studies show how spatial interaction in XR can support
reasoning about abstract data~\cite{marriott2018immersive}, and XR is widely
used in cultural-heritage and museum contexts~\cite{bekele2018arvrmr} for
situating artifacts in space and time. \textit{TimeLineXR} draws on both: the time axis is easier to read when explored in space,
and users can place any image on it (vehicles, instruments, ships, mobile phones, and beyond), which makes it a good fit for museum-style exploration of how objects evolved over time. To our knowledge, no prior demonstration combines a VLM-derived time axis, live image placement, and WebXR interaction in a single package.

\section{System Overview}

\begin{figure*}[t]
\centering
\includegraphics[width=0.74\textwidth]{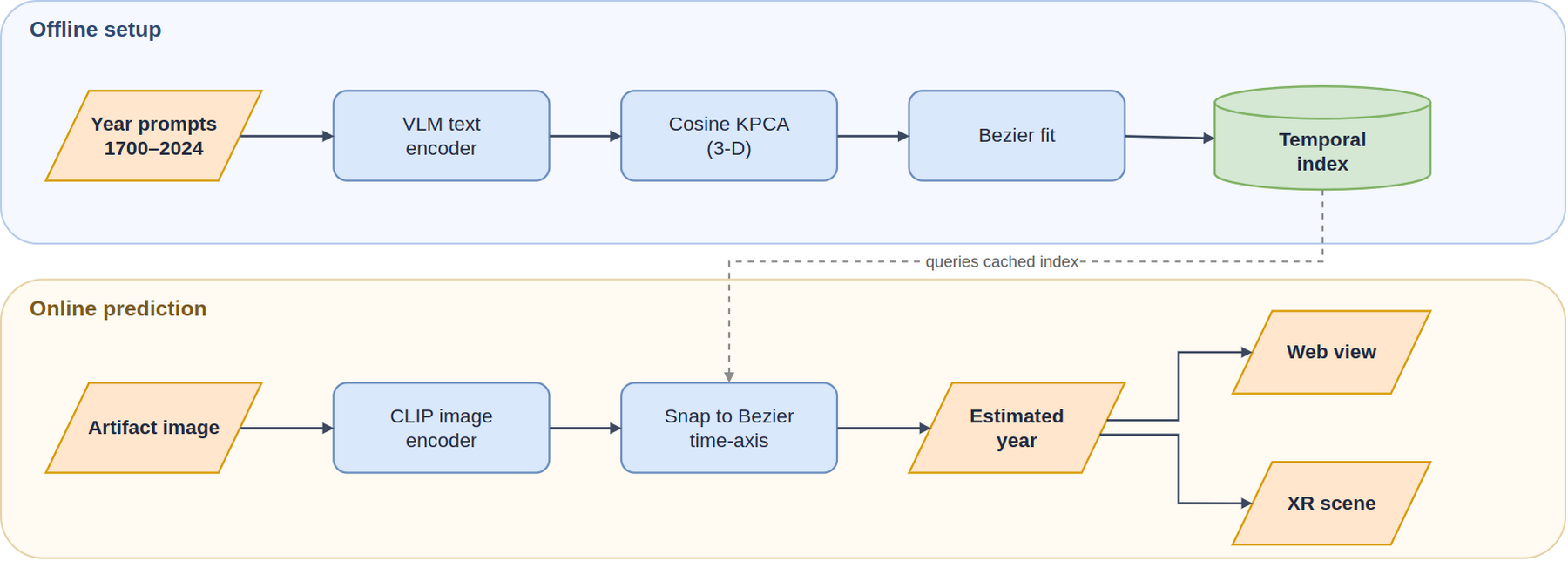}
\caption{System architecture in two phases. The \emph{offline setup} encodes
text prompts for each year, reduces them to three dimensions with cosine-kernel KPCA,
fits a B\'{e}zier curve through the reduced time embeddings, and projects the
embeddings onto the curve to obtain a uniform time axis, caching the result as a
single temporal index. The \emph{online prediction} encodes the query
image with CLIP, snaps it to the B\'{e}zier time axis using the cached
index, and returns an estimated year that the web view and the XR scene
display in their own way. Parallelograms denote inputs and outputs,
rounded rectangles denote processes, and the cylinder denotes the cached
temporal index.}
\label{fig:arch}
\end{figure*}

TimeLineXR builds the time axis offline and then handles queries online
(Fig.~\ref{fig:arch}). The fitted B\'{e}zier time axis and KPCA projection state are computed
once and reused for every query. Users submit a query either through the web upload
controls or by throwing a card in the WebXR scene; both clients run the same estimation
pipeline and return an estimated year together with a 3D coordinate on the time axis,
which the active client uses to place the image in the rendered scene. The time axis
points are the per-year text embeddings, reduced by KPCA and projected onto the
fitted curve, so a placed image lands next to the years that best match it in the
embedding space.

\subsection{Time Estimation Backend}

The time axis is extracted from the VLM following Tekaya et
al.~\cite{tekaya2025matter}: text prompts for each year are encoded by the VLM's
text encoder and reduced to a low-dimensional subspace via cosine-kernel KPCA.
A smooth 1D B\'{e}zier curve is then fitted through these reduced time embeddings,
and each embedding is projected onto its closest point on the curve, so the years
are arranged uniformly along it from 1700 to 2024. At query time, the image is
encoded by the same VLM, projected into the same subspace, and assigned the year
of the closest point on the curve. Figure~\ref{fig:arch} illustrates this pipeline.

For TimeLineXR we reduce to a 3D projection subspace instead of the
13-dimensional subspace of the reference work, so the time axis can be rendered
directly as a navigable 3D scene. Reducing to 3 dimensions increases MAE
compared to the reference setting, as shown in Table~\ref{tab:dim-tradeoff};
this is expected and acceptable since spatial legibility takes priority over
benchmark accuracy in an interactive exploration tool. Users can switch between
CLIP\,ViT-B/32~\cite{radford2021clip} and EVA-CLIP~\cite{sun2023evaclip} from
the web demo to observe how the choice of backbone affects the estimated year.

\begin{table}[ht]
\centering
\caption{Dimensionality trade-off on TIME10k: MAE (years) at $S{=}13$
(reference) vs.\ $S{=}3$ (TimeLineXR), B\'{e}zier($\mathbf{R}^{S}$, NN)
variant of~\cite{tekaya2025matter}.}
\label{tab:dim-tradeoff}
\begin{tabular}{lcc}
\toprule
Backbone & $S=3$ (demo) & $S=13$ (reference) \\
\midrule
CLIP\,ViT-B/32 & $15.30$ & $9.00$ \\
EVA-CLIP       & $20.81$ & $7.62$ \\
\bottomrule
\end{tabular}
\end{table}

\subsection{Web Demo}
\begin{figure*}[t]
\centering
\includegraphics[width=0.92\textwidth]{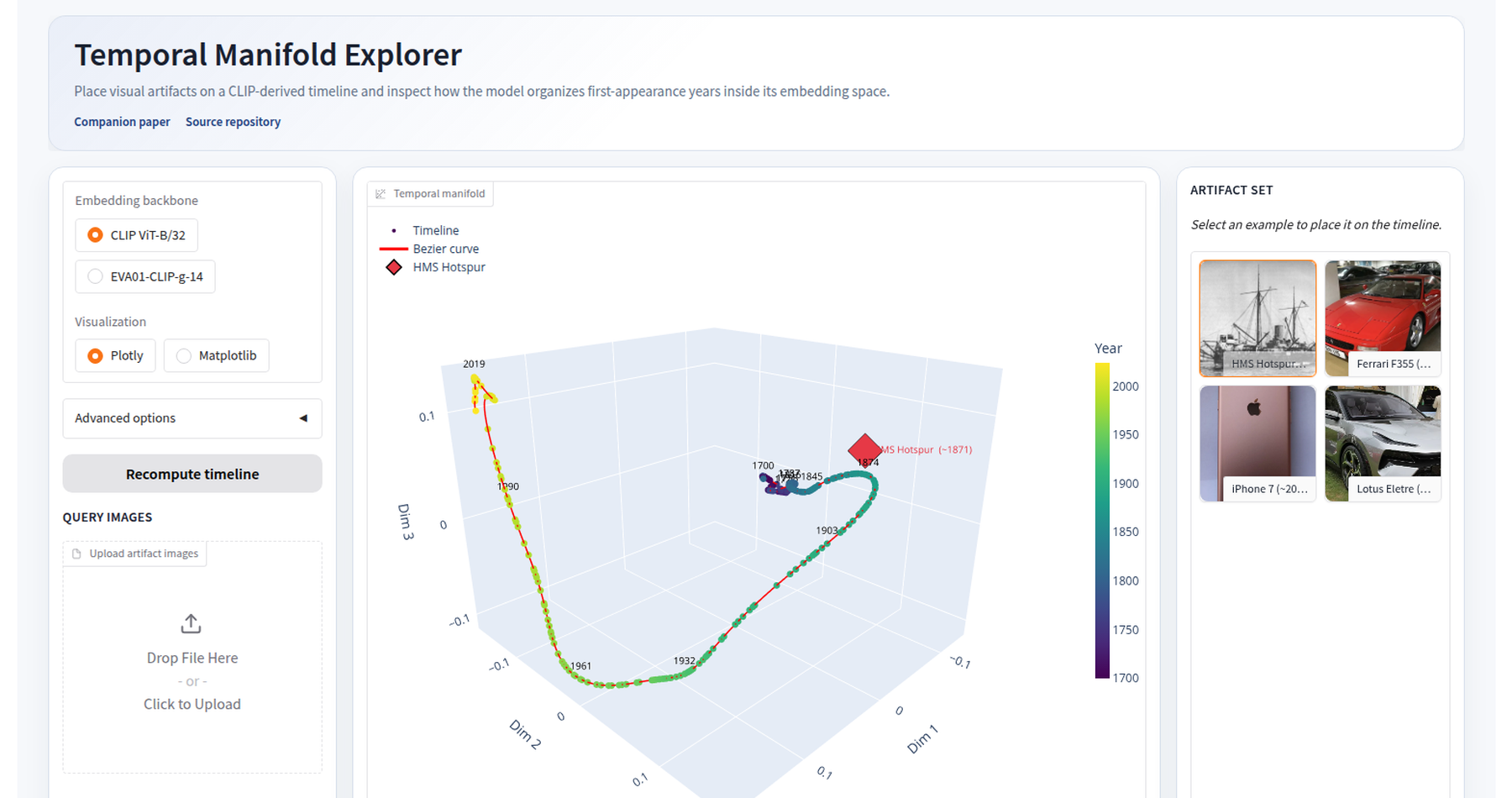}
\caption{Browser web demo after selecting an example. The left panel
specifies the embedding backbone and query path. The centre panel shows the VLM's
time axis in 3D: the time embeddings projected onto the fitted B\'{e}zier curve
(drawn in red) as points colour-coded by year, with the query image placed on it. The
right panel provides bundled examples.}
\label{fig:webdemo}
\end{figure*}

The web demo is a Gradio application that renders the time axis as an
interactive Plotly 3D scatter (Fig.~\ref{fig:webdemo}). Users upload an
image or select a bundled example; the estimated year is added to
the plot and to a running prediction table. The time embeddings, reduced to 3D and
projected onto the fitted B\'{e}zier curve, are shown as points colour-coded by year
on a viridis ramp; the curve itself is drawn as a continuous red line. The demo is a
hosted web application reachable from any browser and is the hardware-free entry point
for users without a headset.

\subsection{WebXR Application}

The immersive frontend is built with Three.js and the WebXR Device API and runs
unmodified in any WebXR-capable browser. Desktop browsers get an orbitable preview;
headsets enter an immersive session that first requests passthrough mixed reality
(\texttt{immersive-ar}) and falls back to a dark virtual environment
(\texttt{immersive-vr}). The time axis is rendered as a room-scale point cloud with a
B\'{e}zier time axis (a tube mesh) running through it. The entire scene can be grabbed,
moved, and pinch-scaled. To query it, the user takes a
card from a grabbable example shelf (or uploads a personal image) and throws it into
the scene. The throw animation is designed to overlap with the backend prediction time: the
animation lasts 0.5--1.3\,s (see Table~\ref{tab:latency}), which is enforced to be
long enough to cover typical prediction times on both CPU and GPU hosts, so the visitor
never sees an idle wait. Beyond masking latency, the motion gives the interaction a
physical feel, as though the user is placing the image into the time axis by hand.
The card then settles at the estimated year on the time axis. The supplementary video
was recorded on a Meta Quest 3 running the Meta Quest Browser, which is also the
device we will use for on-site presentation. Uploaded and bundled images are the
universal input path, since raw passthrough-camera access is not reliably exposed by
current headset browsers.

\section{Demonstration Scenario}

The temporal structure reported in~\cite{tekaya2025matter} exists only as numbers:
correlation scores and error metrics that summarise the time axis but never show its
shape, its dense and sparse regions, or how images attach to it. TimeLineXR makes
that structure directly inspectable, without aiming to improve estimation accuracy.

Users can start from either interface. In the web demo, the time axis begins empty
and images are added one at a time. Adding mobile phones across several decades, the
user watches where each attaches and whether the ordering matches expectation. Adding
objects from a different category over the same period reveals whether the model
groups images by era or by domain, a qualitative observation that aggregate metrics
cannot convey. As images accumulate the global shape becomes visible; at any point
the user can also load a pre-computed time axis from TIME10k~\cite{tekaya2025matter},
populating it with hundreds of images across all six categories, and then add
their own to probe the model beyond the benchmark.

In the XR version on a Meta Quest 3, the same structure is explored at room scale.
Walking along the axis and rotating the point cloud makes the non-linearity tangible
in a way a 2D plot cannot. The demonstration's value is this shift: from a reported
statistic to an inspectable, navigable object.

\section{Implementation Notes}

The backend is implemented in Python with PyTorch, HuggingFace
Transformers, and scikit-learn; it uses frozen pre-trained encoders
and introduces no new training. The web client uses Gradio and Plotly
and will be released as a Hugging Face Space upon acceptance. The
WebXR client is served by a FastAPI/uvicorn application and rendered
in Three.js through a CDN-imported module map, so the only
client-side dependency is a WebXR-capable browser (immersive WebXR
sessions require HTTPS). Table~\ref{tab:latency} reports typical
end-to-end timings for the backend prediction (CLIP encode, KPCA
projection, and B\'{e}zier snap) on a CPU-only host and on a
GPU-enabled host, alongside the XR throw animation duration.

\begin{table}[t]
\centering
\caption{Backend prediction latency and XR throw animation duration.}
\label{tab:latency}
\begin{tabular}{lc}
\toprule
Stage & Time \\
\midrule
Backend prediction (CPU) & $150$--$300$\,ms \\
Backend prediction (GPU) & $<50$\,ms \\
XR throw animation       & $0.5$--$1.3$\,s \\
\bottomrule
\end{tabular}
\end{table}

\section{Future Work}

We see two complementary directions for future work. \textbf{Technical:} Meta Quest and
similar headsets limit what WebXR can access on the device; in particular,
raw passthrough-camera access is not reliably exposed today. Once that changes, we
plan to embed the live camera feed directly into the XR scene so that a
user can point their headset at a physical object and watch its position
appear on the time axis in real time, without an upload step. Lighter on-device backbones such as
MobileCLIP~\cite{vasu2024mobileclip} would make this loop fully local.
\textbf{Applicational:} we plan to deploy TimeLineXR in museums and
exhibitions~\cite{bekele2018arvrmr}, where visitors can follow any chosen
collection (for instance, vehicles, mobile phones, ships, or instruments)
and see how the look of the objects changed decade by decade while
estimating the year of each item in real time.

\section{Conclusion}

TimeLineXR provides an interactive way to explore how a VLM organises time in its
embedding space. We introduced two complementary views, a browser demo and a WebXR
scene, built on a shared estimation pipeline. The demo lets users query the VLM's time
axis with their own images and compare where in time different objects are placed. By
turning a previously statistical finding into an inspectable, navigable object, it
offers a qualitative, complementary view alongside the quantitative results of the
method it builds on.

\section*{Acknowledgements}
This work was funded by the Austrian Science Fund (FWF) under the Visual
Heritage PhD Program doctoral grant [10.55776/DFH37] and by the Austrian
Research Promotion Agency (FFG) under project TrustAI (no. 898085).

\bibliographystyle{IEEEtran}
\bibliography{references}

\end{document}